\documentclass{llncs}

\usepackage{booktabs} 
\usepackage{subdepth}
\usepackage{bm}
\usepackage{dsfont}
\usepackage{xspace}
\usepackage{amsmath}
\usepackage{amsfonts}
\usepackage{graphicx}
\usepackage{cleveref}
\usepackage{pdfpages}
\usepackage[noend]{algpseudocode}
\usepackage{todonotes}
\usepackage{paralist}
\usepackage{pgf}
\usepackage{subfigure}
\usepackage{booktabs}
\usepackage[lined, algoruled, algosection, linesnumbered, resetcount]{algorithm2e}
\usepackage{algpseudocode}

\begin{document}
\title{Improving NeuroEvolution Efficiency by Surrogate Model-based Optimization with Phenotypic Distance Kernels\thanks{
The final authenticated version of this publication will appear in the proceedings of the Applications of Evolutionary Computation - 22nd International Conference EvoApplications 2019 in the LNCS by Springer}}

\author{J\"org Stork, Martin Zaefferer, and Thomas Bartz-Beielstein}

\institute{
TH K\"oln, Institute for Data Science, Engineering, and Analytics\\
Steinm\"ullerallee 1,~51643~Gummersbach, Germany \\
\email{joerg.stork|martin.zaefferer|thomas.bartz-beielstein\\@th-koeln.de}}

\maketitle

\begin{abstract}
In NeuroEvolution, the topologies of artificial neural networks are optimized with evolutionary algorithms to solve tasks in data regression, data classification, or reinforcement learning.
One downside of NeuroEvolution is the large amount of necessary fitness evaluations, which might render it inefficient for tasks with expensive evaluations, such as real-time learning.
For these expensive optimization tasks, surrogate model-based optimization is frequently applied as it features a good evaluation efficiency. 
While a combination of both procedures appears as a valuable solution, the definition of adequate distance measures for the surrogate modeling process is difficult.
In this study, we will extend cartesian genetic programming of artificial neural networks by the use of surrogate model-based optimization. 
We propose different distance measures and test our algorithm on a replicable benchmark task.
The results indicate that we can significantly increase the evaluation efficiency and that a phenotypic distance, which is based on the behavior of the associated neural networks, is most promising.
\end{abstract}

\section{Introduction}
Artificial Neural Networks (ANN) are utilized in many different fields, such as data regression, data classification, or reinforcement learning. 
In each of these tasks, the network topology is significant for the ANNs performance. 
Often, only parameters such as the edge weights, number of hidden layers, and number of elements
per layer are considered during the optimization of an ANN.
In NeuroEvolution (NE), ANNs are generated by Evolutionary Algorithms (EAs). 
NeuroEvolution allows severe modifications of networks, 
such as individual connections between neurons and different neuron transfer functions, 
leading to a large search space of potential topologies. 
Two example algorithms of this category are NeuroEvolution of Augmenting 
Topologies (NEAT)~\cite{stanley2002evolving} 
and Cartesian Genetic Programming of Artificial Neural Networks (CGPANN)~\cite{miller2000cartesian,turner2013cartesian}. 
Both NEAT and CGPANN require numerous fitness evaluations to find adequate topologies in the large search space. 
This makes them inefficient for applications with expensive function evaluations such as simulations or real-world
experiments, e.g., when an ANN is the controller of a robot that operates in a complex real-time environment.
Surrogate Model-based Optimization (SMBO) is frequently used for data efficient optimization of real-world processes, as it has a high evaluation efficiency~\cite{koziel2013surrogate}. 
Surrogate models are commonly employed in continuous optimization, where Euclidean spaces and distance metrics exist. 
The more complex discrete search spaces are less often investigated~\cite{Bart16n,Zaefferer2014b}.
In this work, we want to extend CGPANN to employ Surrogate Model-based NeuroEvolution (SMB-NE) to reduce the load of fitness evaluations. 
For this task, we defined several distances based on the CGPANN genotypes and a distance that measures, instead of the structure, the difference in behavior of the ANNs. 
Our approach allows the definition of a  phenotypic distance that is indifferent to the size or topology of an ANN. 
It only requires a representative input set to model the input to output correlation.
Our main research questions are:
\begin{enumerate}
\item Is SMB-NE able to outperform CGP NeuroEvolution in terms of evaluation efficiency without loss of accuracy?
\item How can we create a representative input set for the phenotypic distance in SMB-NE?
\end{enumerate}
For all experiments, we utilize data-mining classification tasks as a controllable and cheap to evaluate benchmark for the learning efficiency of SMB-NE using few fitness evaluations.
This article is structured as follows: Section~\ref{sec:rel} discusses related work. 
Section \ref{sec:smbne} illustrates the utilized methods and the SMB-NE algorithm. 
Section \ref{sec:kernels} introduces the different distance measures for SMB-NE. 
In Section \ref{sec:exp} the experimental setup and the results are shown and further discussed. 
Finally, in Section \ref{sec:con} we conclude the paper and give an outlook to future work. 

\section{Related Work}\label{sec:rel}
This work is an extension of an unpublished study presented at an informal workshop.\footnote{For sake of the double blind review process, we did not present references/or the name of the workshop.}
A recent study by Zaefferer et al.~\cite{Zaef18b}  investigated the use of SMBO in Genetic Programming with genotypic and phenotypic distance measures.
In constrast to this study, they performed tests comparing an EA with SMBO for a bi-level symbolic regression optimization problem. 
They concluded, that the SMBO approach is able to outperform the EA in terms of evaluation efficiency. 
Moreover, the phenotypic distance performed best and was very fast to compute.
A first surrogate model for ANN optimization was applied to NEAT by Gaier et al.~\cite{gaier2018data}.
They use a surrogate-assisted optimization approach by combining an EA with a surrogate distance-based model, employing a genotypic compatibility distance that is part of NEAT.
With this approach, they are also able to increase the evaluation efficiency. 
To our best knowledge, nobody else used a surrogate model-based approach for the task of NeuroEvolution. 
Phenotypic distances are also used in other applications, for example, the optimization of job dispatching rules. 
For this tasks, Hildebrandt and Branke~\cite{Hildebrandt2014} utilize a surrogate-assisted EA, which is able to increase 
the evolution towards good solutions.
The use of surrogates including genotypic distances is more often discussed, e.g., for optimization of fixed neural network topologies in reinforcement learning \cite{Stor17c}.

\section{Data Efficient Optimization for Neuroevolution} \label{sec:smbne}
\subsection{Neuroevolution by Cartesian Genetic Programming}
\begin{figure}[ht]
\centering
\includegraphics[width=0.9\textwidth]{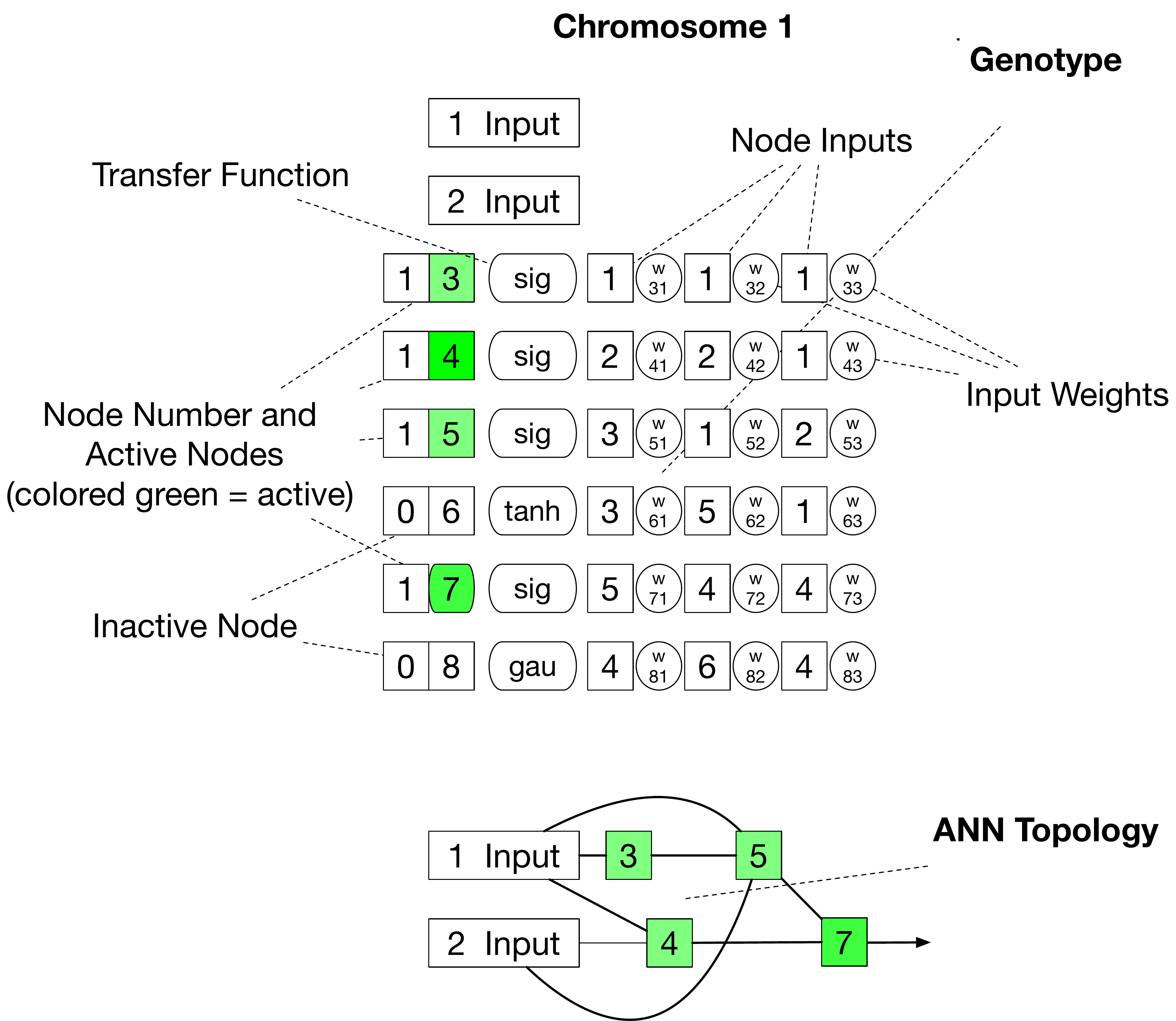}
\caption{A CGPANN genotype with two inputs, eight nodes, an arity of three and different transfer functions. Each node has a transfer function, a boolean activity gene and several inputs with adjacent weights. Green nodes are active and part of the encoded ANN. }
\label{fig:chromosome}
\end{figure}
For NeuroEvolution with CGP, we use the C library \texttt{CGP} by A.~Turner\footnote{http://www.cgplibrary.co.uk - accessed: 2018-01-12}, which also allows the application of NeuroEvolution~\cite{turner2015introducing}. 
The \texttt{CGP} library was modified with function interfaces to the statistical programming language R, distance measures described below, and additional fitness functions. 
The genotype of a CGPANN individual consists of a fixed number of nodes. 
Each node has a number of connection genes based on the pre-defined arity with adjacent weight genes and a single categorical function gene. 
Nodes are only connected to preceding nodes. Duplicate connections to nodes are possible and if present, adjacent weights will be added to form a single connection in the resulting ANN. 
Moreover, each node has a Boolean activity gene, which signals if it is used in the active ANN topology. 
Fig.~\ref{fig:chromosome} displays an example for a CGPANN genotype and the encoded ANN with two inputs, 
eight nodes and an arity of three. 
In CGP NeuroEvolution, the network topology and weights are optimized using an evolutionary approach, utilizing a (1+4)-Evolution Strategy (ES), i.e., one parent and four new individuals in each generation. 
In contrast to the standard selection in ES, the rank-based selection process favors the offspring, and thus the new solution, over a parent with the same fitness. 
Different mutation operators are available, the default is probabilistic random mutation. 
Inactive genes will not influence the fitness of an ANN.

\subsection{Kriging}
Our SMBO algorithm is based on a Kriging regression model, 
which assumes that the observations are samples from a Gaussian process~\cite{Forrester2008a}.
Kriging is a kernel-based model, i.e., the model uses a kernel, or correlation function,
to measure the similarity of two samples.
One typical kernel for real-valued samples is the exponential kernel, i.e., $\text{k}(x,x')=\exp(-\theta ||x-x'||_2)$.
Here, the kernel parameter $\theta$ expresses how quickly the correlation decays to zero, with increasing Euclidean distance $||x-x'||_2$ between the samples. The parameter is determined by Maximum Likelihood Estimation (MLE).
It is straightforward to extend kernel-based models 
to combinatorial search spaces~\cite{Moraglio2011,Zaefferer2014b}. 
Essentially, the Euclidean distance is replaced by a corresponding distance that applies to the
respective search space, with the exponential kernel $\text{k}(x,x')=\exp(-\theta \text{d}(x,x'))$. 
The design of appropriate distances $\text{d}(x,x')$ for neural networks is in the focus of this paper.

Kriging is frequently employed in SMBO algorithms, because in addition to relatively accurate predictions,
it also provides an estimate of the uncertainty of each prediction.
The predicted value and the uncertainty estimate are used to compute infill criteria. 
These infill criteria are supposed to express how desirable it is to evaluate a new candidate solution $x$.
To that end, the uncertainty estimate is integrated to push away from known, well-explored areas, 
instead preferring solutions that have large uncertainties, yet promising predicted values.
One of the most frequently used criteria is Expected Improvement (EI)~\cite{Mockus1978,Jones1998}.
We employ the Kriging implementation of the R package 
\texttt{CEGO}~\cite{CEGOv2.2.0,Zaefferer2014b}.
It uses distance-based kernels to model data from structured, combinatorial search spaces.

\subsection{Surrogate Model-based NeuroEvolution (SMB-NE)}
\begin{algorithm}
 \caption{Surrogate Model-based Neuroevolution}
 \label{alg:smdne} 
 \Begin{
\tcp{phase 1: initialization} 
 $t=1$ \\
 \textbf{initialize} $k$ neural networks ($x_i$) at random  \\
 \textbf{evaluate} their fitness on the objective function \\
 \textbf{build} Kriging surrogate model utilizing $s_{t}(x_i)$ and distance measure $d(x_i,x_j)$;   \\
 \tcp{phase 2: optimization} 
 \While{\textbf{not} termination-condition}{
\If{$t > 1$}{\textbf{rebuild} surrogate model $s_{t}$ with a set ${\cal{M_t}}_{t} \in {\cal{D}}_{t}$ of observations }
\textbf{optimize} EI with a (1+4)-ES to discover improved $x_{t}$  \\
\textbf{evaluate} network $x_{t}$ fitness $y_{t}$ on the objective function \\
\textbf{add} evaluated networks to archive ${\cal{D}}_{t+1}= \{{\cal{D}}_{t},(x_{t},y_{t})\}$  \\
 $t=t+1$
 }
}
\end{algorithm}
We extend CGP NeuroEvolution with an SMBO approach, leading to the Surrogate Model-based NeuroEvolution (SMB-NE) strategy outlined below. 
The strategy is intended to perform a data efficient search by predicting the fitness of candidate solutions with the help of a Kriging surrogate model. 
The algorithm is outlined in \ref{alg:smdne}.

The SMB-NE process starts by creating a random initial set of individuals, in our case ANNs, and 
evaluates them with the objective function.
The resulting data is used to learn a Kriging model.
For learning the model, we define different genotypic and phenotypic distance measures for ANNs in Sec.~\ref{sec:kernels}. 
With the model, we are able to estimate the EI of an individual.
In each following iteration, the model is utilized to suggest new promising ANNs
by optimizing the EI criterion with the (1+4)-ES algorithm of CGP.
The (1+4)-ES is used to introduce the ability to directly compare CGPANN to SMB-NE, without an additional optimization algorithm with a different operator set, or parametrization influencing the results.
In each iteration, the single most promising individual is evaluated and added to the archive ${\cal{D}}_{t}$.
As the archive grows during the optimization process, the computational effort of creating the Kriging model rises with $O(m^3)$, where $m$ is the surrogate model sample size.
To keep the computational effort on a feasible level, a subset ${\cal{M}}_t \in {\cal{D}}_{t}$ of size $m$ is used for the modeling process.

This modeling set ${\cal{M}}_t$ is formed by selecting $\frac{m}{5}$ of the best and $\frac{4*m}{5}$ randomly drawn individuals out of the archive in each iteration. 
We chose the fractions in the strategy to ensure a balance between exploration and exploitation.
If the size of the archive ${\cal{D}}_{t}$ is smaller than or equal to the size of ${\cal{M}}_t$, all individuals are selected. The influence of the size of the modeling set ${\cal{M}}_t$ is investigated in Sec.~\ref{sec:modelsize}.

\section{Proposed Kernels and Distances}\label{sec:kernels}
In the following, we always use an exponential kernel $k(x,x')=\exp(-\theta d(x,x'))$.
Here, $x$ and $x'$ represent ANNs. They consist of 
the weights $x_{w}$, input labels $x_i$, activity labels $x_a$, and transfer function labels $x_f$, i.e.,
$x=\{x_w,x_i,x_a,x_f\}$.
All distances $d(x,x')$ are scaled to $[0,1]$.  
In addition, we also considered employing the graph edit distance, 
but decided against it due to its complexity. Computing the graph edit distance is NP-hard~\cite{Zeng2009}.

We illustrate the different distances by providing an example that compares two specific ANNs, with a similar structure as outlined in Fig.~\ref{fig:chromosome}. 
For the sake of simplicity and comparability, all weights are set to 1.
The two ANNs only differ with respect to two nodes. 
The transfer function and connections of one active (node 3) and one inactive node (node 8) are changed.

\subsection{Genotypic Distance (GD)} 
\begin{figure}[h]
\centering
\includegraphics[width=0.85\textwidth]{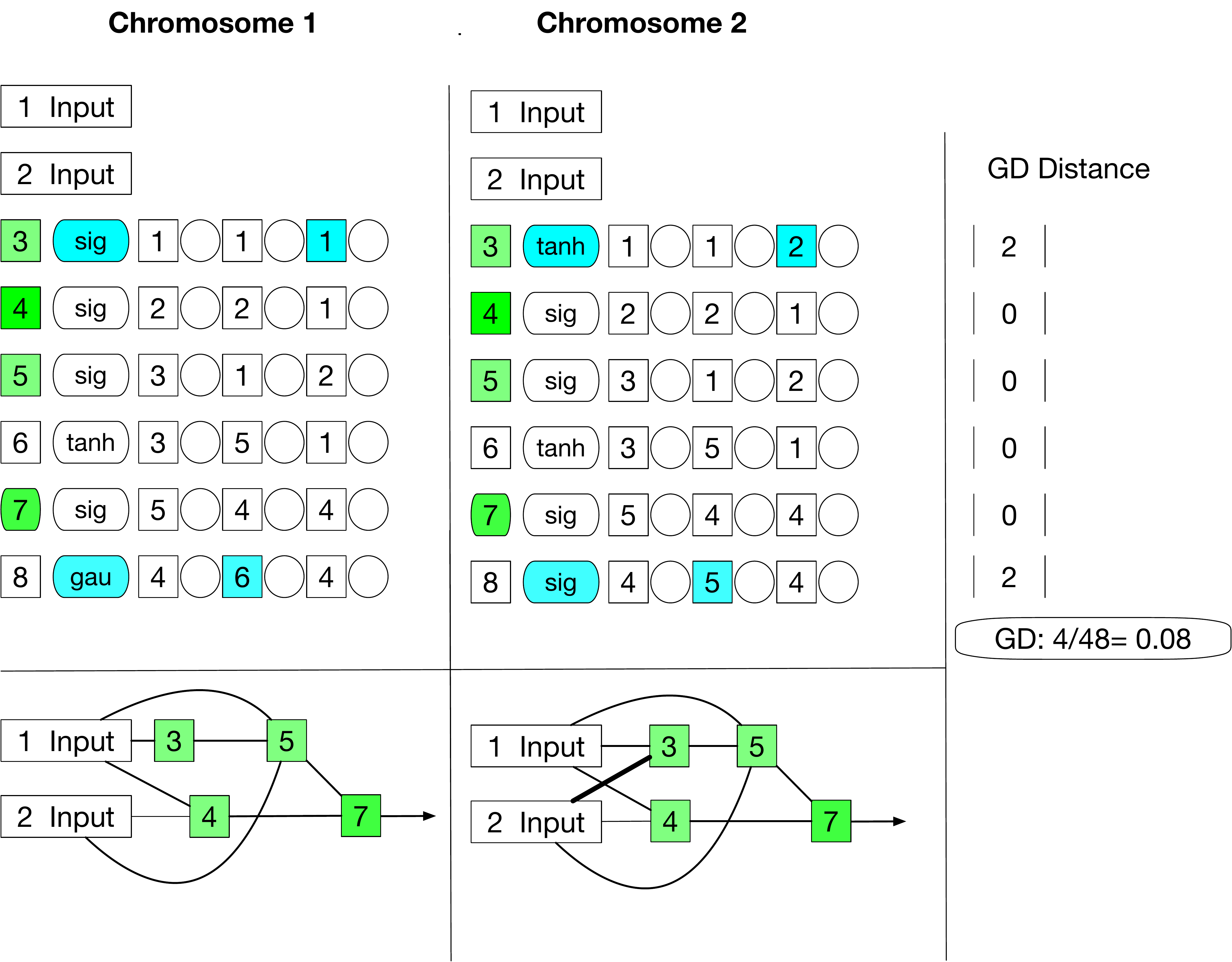}
\caption{Example calculation the GD distance for two distances. By introducing only small changes to the genotype, the normalized GD stays rather small.}
\label{fig:gd}
\end{figure}
The genotypic distance is based solely on the genotype of a CGPANN individual. 
As the CGPANN genotypes have a fixed structure, the distance of two individuals can be calculated by a row-wise comparison of their nodes. By  combining the distances of weights, inputs, activity of nodes, and transfer functions we obtain a distance  $d(x,x')= ||x_w-x_w'||^2_2+ H(x_i,x_i') +H(x_a,x_a') +H(x_f,x_f')$, where $H(a,b)$ denotes the Hamming distance, i.e., $H(x_f,x_f')$ is 0 if the transfer function is identical, else $H(x_f,x_f')$ is 1. 
The GD is further normalized by the total number of possible comparisons for the given genotypes. 
Fig. \ref{fig:gd} illustrates an example for calculating the GD distance. 
Although taking the non active nodes into account, the normalized GD is rather small.  
We further extended  the GD by ordering the inputs of each ANN before the calculation to match the weight distances to the correct input if (and only if) two nodes have similar connections, but a different ordering of inputs in the genotype.

\subsection{Genotypic ID Distance (GIDD)} 
\begin{figure}[h]
\centering
\includegraphics[width=1\textwidth]{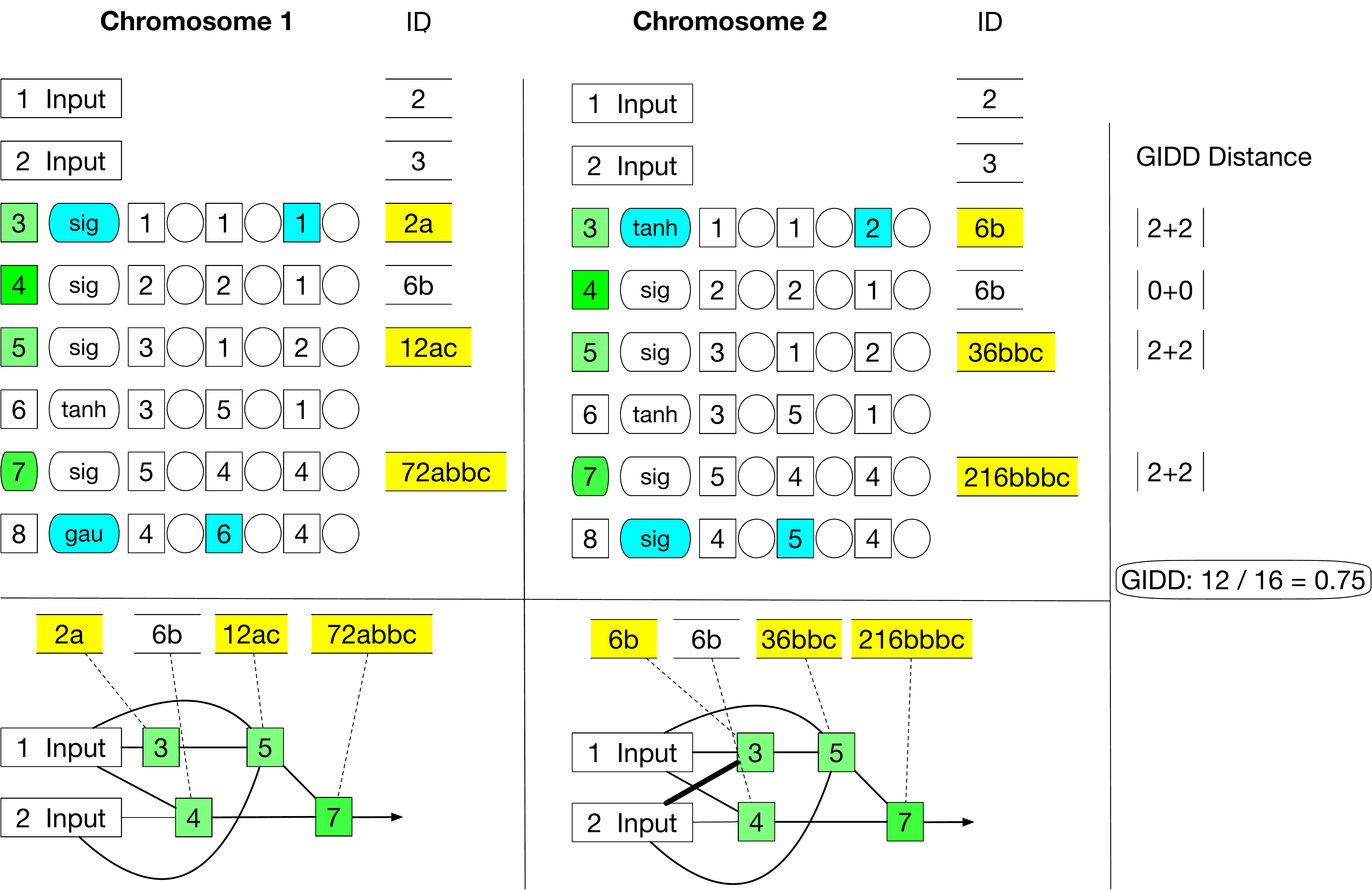}
\caption{Example calculation for GIDD. The IDs are based on prior nodes or inputs and the number (a=1, b=2, c=3) of non-duplicate connections. The calculated distance is high because of the different node IDs, which are based on their relative position in the ANN.  }
\label{fig:gidd}
\end{figure}
The GIDD is intended to solve an important issue of the genotypic distance: different nodes and adjacent functions and weights in one row of a CGPANN genotype are not easily comparable, if their influence on the resulting ANN and also phenotype is considered.
The idea behind the GIDD is to only compare those nodes, which are placed in the same position in the ANN and to solve the problem posed by competing conventions, i.e., that a certain ANN topology can be expressed by numerous genotypes.  
The distance is based on the active topology and creates IDs which are designed to be unique for an equal placement of a node in the ANN. 
Inactive nodes do not influence the GIDD. 
Thus, each active node in the ANN is given an ID based on the connections to prior nodes or inputs and the number of non-duplicate 
connections of this node. Then, the distance of nodes can be calculated by a pairwise comparison of all node IDs.
If a certain node ID matches for both ANNs, the subgraph prior to this node is analyzed recursively for validation of the ANN up to the position of the matching nodes. 
If all IDs in the subgraph are identical, we assume that the corresponding nodes have an equal position in the ANN topology. 
For all nodes that are matched in this way, the Euclidean distance of the weights ($x_w$) and Hamming distance of
the transfer functions ($x_f$) is computed.
A node pair can only be used once for this comparison, as node IDs may be present several times in each individual. 
If all node IDs of both individuals $x$ and $x'$ are equal, the GIDD is simply the distance $d(x,x')= ||x_w-x_w'||^2_2 +H(x_f,x_f')$ between
all weights and transfer functions. 
If nodes do not match, for each node not present in the compared ANN, a fixed distance is assigned (in our example 2). 
Again, the GIDD is normalized by the maximum distance of two individuals.  
Fig. \ref{fig:gidd} illustrates the calculation in an example.
Contrary to the GD, the GIDD reacts strongly to the introduced changes, as they have a large influence on the node relations, which results in different node IDs and puts them to maximum distance. 

\subsection{Phenotypic Distance (PhD)} 
\begin{figure}[h]
\centering
\includegraphics[width=1\textwidth]{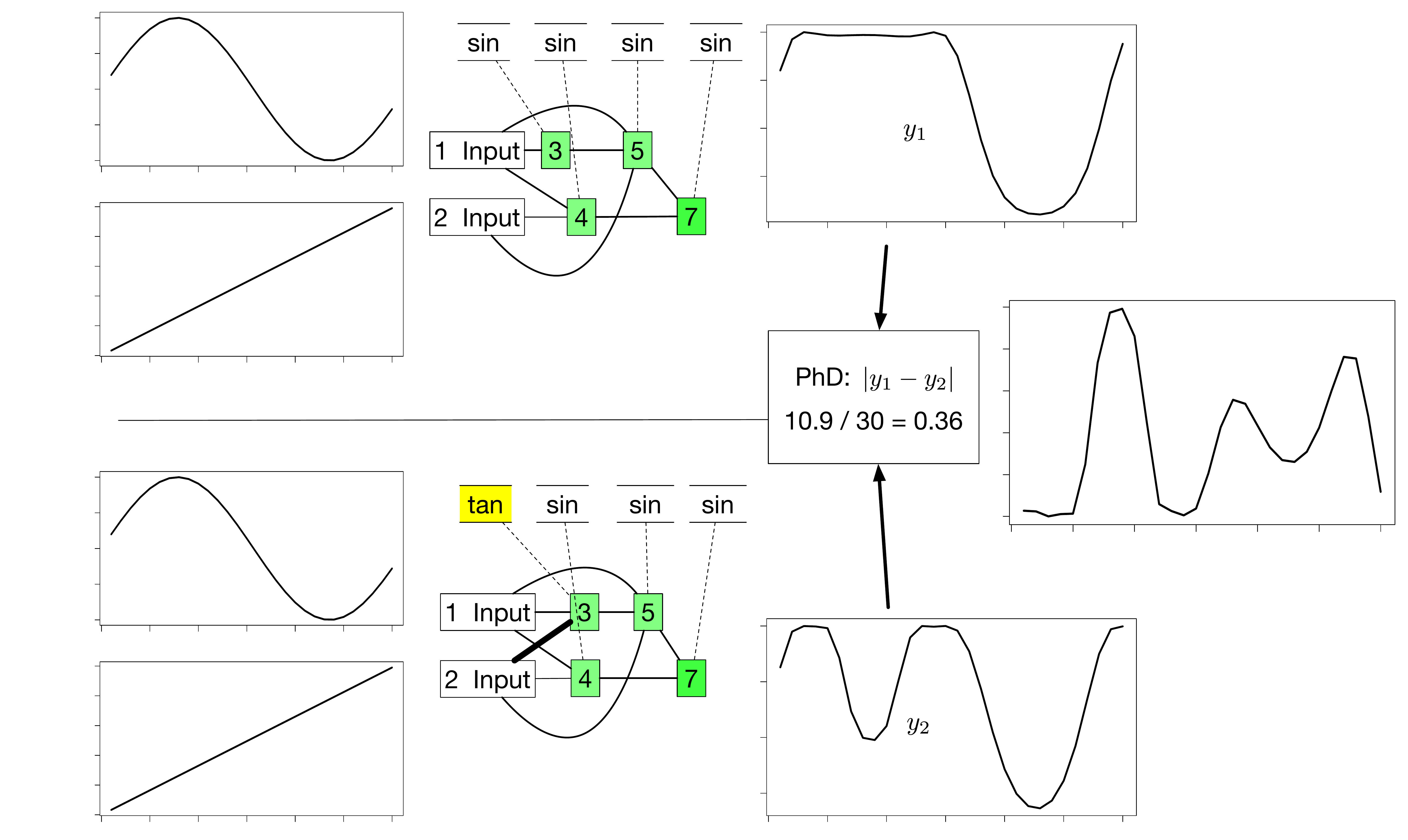}
\caption{PhD example with continuous inputs and trigonometric transfer functions. Two input samples, from a sine and linear function of length 30 are fed to two ANNs, which differ in their transfer functions and a single connection. The phenotypic distance is the normalized absolute difference of their output signals.}
\label{fig:phd}
\end{figure}
The phenotypic distance does not utilize any genotypic information of the ANNs to compute the distance. 
Instead, it utilizes solely their behavior.
In our definition, the phenotype of a neural network is how it reacts to a certain input set, i.e., which outputs are produced. 
This output is then compared to resolve in a distance measure, which is indifferent to changes to the underlying genotype, including transfer functions, weights or connections, which result in the same behavior. 
More importantly, it is insensitive to the size of the genotype.
The PhD distance utilizes the L1 norm distance to account for large dimensions of the input and output vector and is further normalized by the input set length. 
While it is indifferent to the network topology in terms of encoding, size, number of connections etc., it is very sensitive to the choice of the input set. 
Thus, these input sets have to be carefully selected to be representative for the underlying task and/or environment. 
In Section \ref{sec:doe} we examine the influence of the input set for the task of classification.
An example calculation for the distance of small ANNs utilizing PhD is given in Fig. \ref{fig:phd}.
In the example, two continuous input samples, from a sine and linear function of length 30 are fed to two ANNs, where they differ in their transfer functions and a single connection. 
As this example is intended for understanding the change in the output signals, simple trigonometric functions have been used, which are commonly utilized in Genetic Programming. 
The small example shows that minor changes in the network topology can result in a large difference in the phenotype.

\subsection{Mixed Distance (MD)} 
Similar to linear combination distance in \cite{Zaef18b}, we utilize a mixed distance of GD, GIDD, and PhD, where each distance receives a weight $\beta_i \in \mathbb{R}^+$ determined by MLE. As the performance of each distance is unknown a-priori, the idea behind the MD is that allows an automatic selection of the most adequate distance measure in each optimization step. 

\section{Experiments} \label{sec:exp}
To assess the ability of SMB-NE to improve the efficiency of optimizing the topologies of ANNs, we decided to perform experiments with classification tasks. 
Classification is well understood, easily replicable, and does not introduce complex problems with the selection of environments or tasks as in general learning (e.g. reinforcement learning).
We limited the experiments to a small budget of function evaluations, which provides a realistic scenario for problems with expensive fitness evaluations, such as real-time learning. 
The experiments are twofold: 
\begin{itemize}
\item First, we estimate the ability of SMB-NE to learn an elementary data set comparing the introduced distance measures GD, GIDD, PhD and MD.
\item Second, we further research how SMB-NE using the PhD reacts to different inputs sets and surrogate model sizes.
\end{itemize}

\subsection{Comparison of Distance Measures for SMB-NE}
\subsubsection{Experimental Setup} 
For the first set of experiments, the well-known IRIS data set is used as an elementary benchmark problem. 
IRIS has $n=150$ samples, 4 variables, and 3 evenly distributed ($n=50$ for each) classes of different flower types (Iris setosa, Iris virginica and Iris versicolor).
The focus of this benchmark is the capability of SMB-NE to learn the best network topology to classify the data set with only 250 fitness evaluations. 
The fitness function is the adjusted classification accuracy: 
$\text{acc}=\sum_{i=1}^n a_i$, where $a_i=1$ if the predicted class is true, otherwise, $a_i$ is the predicted probability for
the true class. 
ANN optimized with Random Search and the inbuilt (1+4)-ES of CGPANN with different mutation rates are being used as baselines.
For SMB-NE, all above described distance measures are compared, while for PhD four different input sample sets are tested. 
As a baseline, we used the complete IRIS data set and additional factorial designs(FD) with small (15) and a large (60) sizes as well as a Latin Hypercube Sample (LHS) with 150 samples. 
Both the FD and LHS are based on the IRIS variable boundaries. 
Further, the output of the PhD is adapted by computing the \emph{softmax}, which yields the class probabilities.
In this experiment, for the (1+4)-ES of CGPANN and SMB-NE pure probabilistic random mutation is used with a strength of  5\%.
In SMB-NE, the (1+4)-ES is used in each consecutive iteration alternating between exploitation (L) and exploration (G). 
Parameters are listed in Table~\ref{tab:setup}. 
The ANNs in this benchmark were kept rather small with a maximum of 40 nodes and 200 connections.
This was intended to keep the search space small, but sufficient for the IRIS data set. 
Algorithm parameters were not tuned and all experiments were replicated 30 times. \\
\begin{table}[h]
\centering
\caption{Parameter setup, where evaluations denote the initial candidates plus the budget for consecutive evaluations. 40 nodes were used to keep the search space small, but sufficient for the IRIS problem given this small budget.}
\label{tab:setup}
\begin{tabular}{l|l|l|l}
  \bf arity & \bf nodes & \bf weight range & \bf function set \\
  5 & 40 &  {[}-1,1{]} & \multicolumn{1}{l}{ tanh, softsign, step, sigmoid, gauss}\\
\midrule
\bf method  & \bf mutation rate & \bf evaluations & \bf surrogate evaluations \\
Random &   & 250 &  \\
CGPANN &  5\%/15\% &  $1+4\cdot63$&   \\
SMBO &  L:5\% G:15\% & 50+200 & L:10+400 G:1000+400\\
\end{tabular}
\end{table}

\subsubsection{Results} 

\begin{figure}[ht]
\centering
\includegraphics[width=1\textwidth]{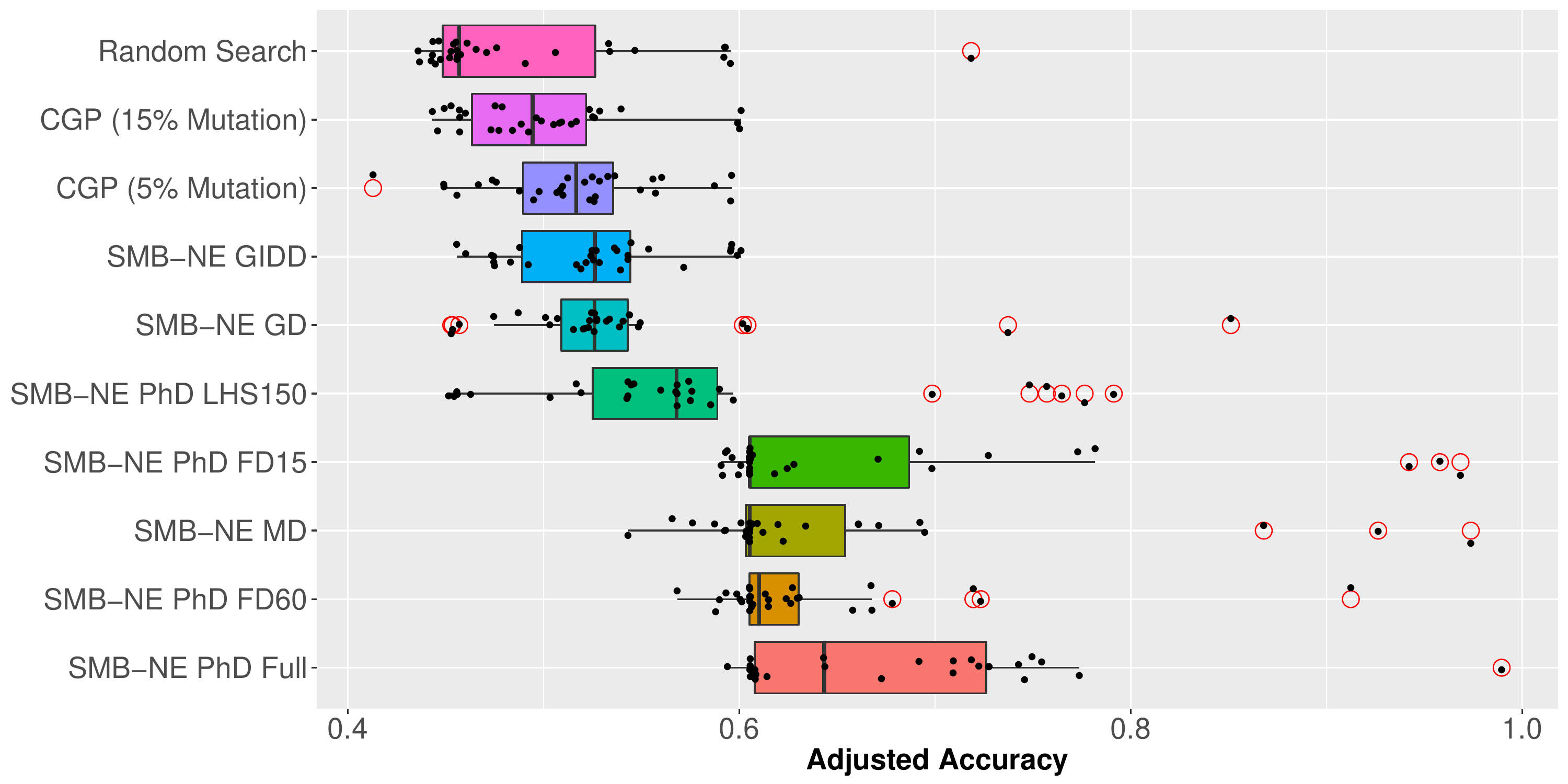}
\caption{Results after 250 fitness evaluations with 30 replications, comparing random search, original CGPANN ((1+4)-ES) and different SMB-NE variants. The results were ranked (top down) by median values. 
Red circles depict outliers.}
\label{fig:results}
\end{figure}

Fig. \ref{fig:results} visualizes the results. 
Firstly, the results show that the standard (1+4)-ES of CGPANN performs better than random search, even with the small number of evaluations. 
A low mutation rate, which depicts a rather local search, seems to be beneficial in this case.
SMB-NE utilizing the GD and GIDD distance measures performs only slightly better than the basic CGPANN. 
With the use of PhD, a significant performance increase is observed, while the PhD with the complete input set performs the best and the LHS the worst.
The mixed distance, which also utilizes PhD with complete input sets, cannot benefit from the linear combination, but is able to deliver a close performance to the sole use of PhD. 
Most runs seem to end up at a local optimum around an accuracy of 66\%, which can be explained by the fact, that at this point two out of three classes of the IRIS data set are predicted correctly.

\subsubsection{Discussion} 

An important insight of this experiment is that in comparison to the PhD, the genotypic distances (GD, GIDD) show a poor performance, even for the small genotype size. This fact might be explained by the small correlation between changes to the genotype and the resulting phenotype: 
small changes to the genes can have a massive impact on the behavior.
Moreover, the GD has the problem that the calculated distances do not consider that aligned row-paired nodes in the genotype can have different, not similar placements in the ANN topology. 
This results in a misleading distance calculation of weights, inputs and transfer function for these nodes. 
However, even the more complex GIDD, which may be able to avoid this issue, does not seem to provide significantly better results. 
Further given the fact that the GIDD is computationally very expensive to calculate and has also shown numerical problems for large genotypes (as it utilizes several recursions over the complete ANN), it is further considered as not suitable for the task of SMB-NE. 
In contrast, the PhD distances show very promising results by directly exploiting the ANN behavior. 
Importantly, the small factorial input sample, which is very fast to compute, shows a good performance. 
Given the poor performance of GD and GIDD, the MD is able to automatically select the PhD distance measure and deliver equal performance.
\noindent
For the second set of experiments, we thus focus on exploiting features of SMB-NE utilizing the PhD distance measure. 

\subsection{Influence of the Input Set and Surrogate Model Size using SMB-NE with PhD}
\label{sec:modelsize}
\label{sec:doe}
\subsubsection{Experimental Setup} 

\begin{table}[h]
\centering
\caption{Algorithm parameter setup for the second set of experiments. The size of the genotypes and the number of evaluations was significantly increased.}
\label{tab:setup2}
\begin{tabular}{l|l|l|l}
  \bf arity & \bf nodes & \bf weight range & \bf function set \\
  25 & 100 &  {[}-1,1{]} & \multicolumn{1}{l}{ tanh, softsign, step, sigmoid, gauss}\\
\midrule
\bf method  & \bf mutation & \bf evaluations & \bf surrogate evaluations \\
CGPANN &  single active &  $1+4\cdot137$ (x10, x100) &   \\
SMB-NE &  single active & 50+500 & L:10+400 G:1000+400\\
\end{tabular}
\end{table}
To assess the performance of PhD, we discarded the GD, GIDD, and MD distance and introduced two more complex data sets, the glass and the cancer data set\footnote{Available in the UCI machine learning repository: https://archive.ics.uci.edu/ml/index.php}. 
Both data sets were preprocessed by normalization and subsampling.
Glass has 9 variables (material composition), 6 unevenly distributed classes of different glass types and 214 samples, while cancer has 9 variables, 2 classes (cancer yes/no) and 699 samples. 
Two benchmarks were conducted. 
The first benchmark investigated the influence of the input sample set used for generating the PhD distance measure. 
For the benchmark, different input samples were created by design of experiments (DOE) methods. 
All DOE sets are based on the known variable ranges and are not subsamples of the original data sets. 
The lengths of the samples are identical for both data sets, as they have the same number of variables.
We compared the following: 
\begin{enumerate}
\item Small and large factorial designs, including main, interaction and quadratic effects, with 55/157 samples each.
\item Small and large Latin Hypercube samples, with 55/110 samples each.
\item The complete datasets as baseline input set, with 214/699 samples each.
\item As algorithm baseline, CGPANN with an increasing number of evaluations ($5.5 \times 100$,  $5.5 \times 1,000$, and $5.5 \times 10,000$).
\end{enumerate}
The motivation of this benchmark is as follows: in real-world optimization tasks, often  \textit{a priori} information of the the task and/or environment is sparse, as the underlying problem is a black-box problem. 
Thus, initially no data set is available to serve as an input set and the user has to rely on design of experiment methods to create input data for the PhD.
\noindent
Further, several changes were made to the algorithm setup to account for the more complex classification problems:
\begin{itemize}
\item The genotype size was significantly enlarged to 100 nodes with 25 arity, resulting in a maximum of 2500 weights/connections.
\item For all compared algorithms, the mutation operator is changed to Goldmans single mutation, which mutates exactly one active node in the genotype (and an arbitrary number of inactive nodes).
\item The number of total function evaluations was raised to 550, while fixing the surrogate model set ${\cal{M}}_t$ size to a value of 100. Each sample hereby consists of 80\% random and 20\% of the most fit  individuals from the solution archive ${\cal{D}}_t$.
\end{itemize}
Table \ref{tab:setup2} shows the adjusted algorithm parameter setup.
In the second, connected benchmark, the influence of the surrogate model size, which is used in each iteration of SMB-NE, to the overall optimization performance is analyzed. 
The glass data set with the complete input set and cancer with the factorial input set are compared for different model set ${\cal{M}}_t$ sizes 

\subsubsection{Results} 
\begin{figure}[!h]
\centering
\includegraphics[width=1\textwidth]{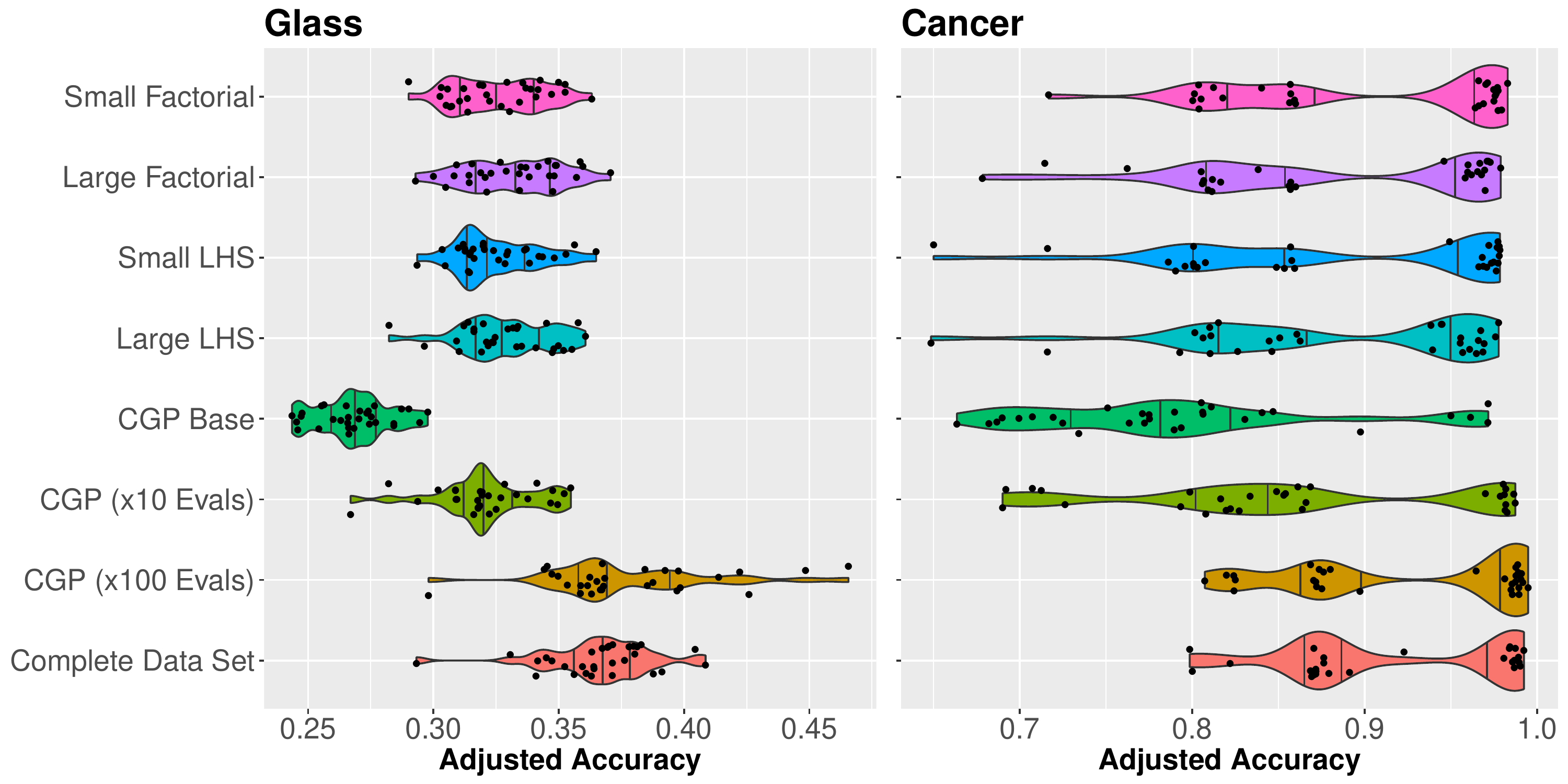}
\caption{Violin plot of the benchmarks with SMB-NE using PhD for different input sets created by experimental design techniques. The results are compared to the complete dataset and CGPANN using a different number of total function evaluations}
\label{fig:resultDOE}
\includegraphics[width=1\textwidth]{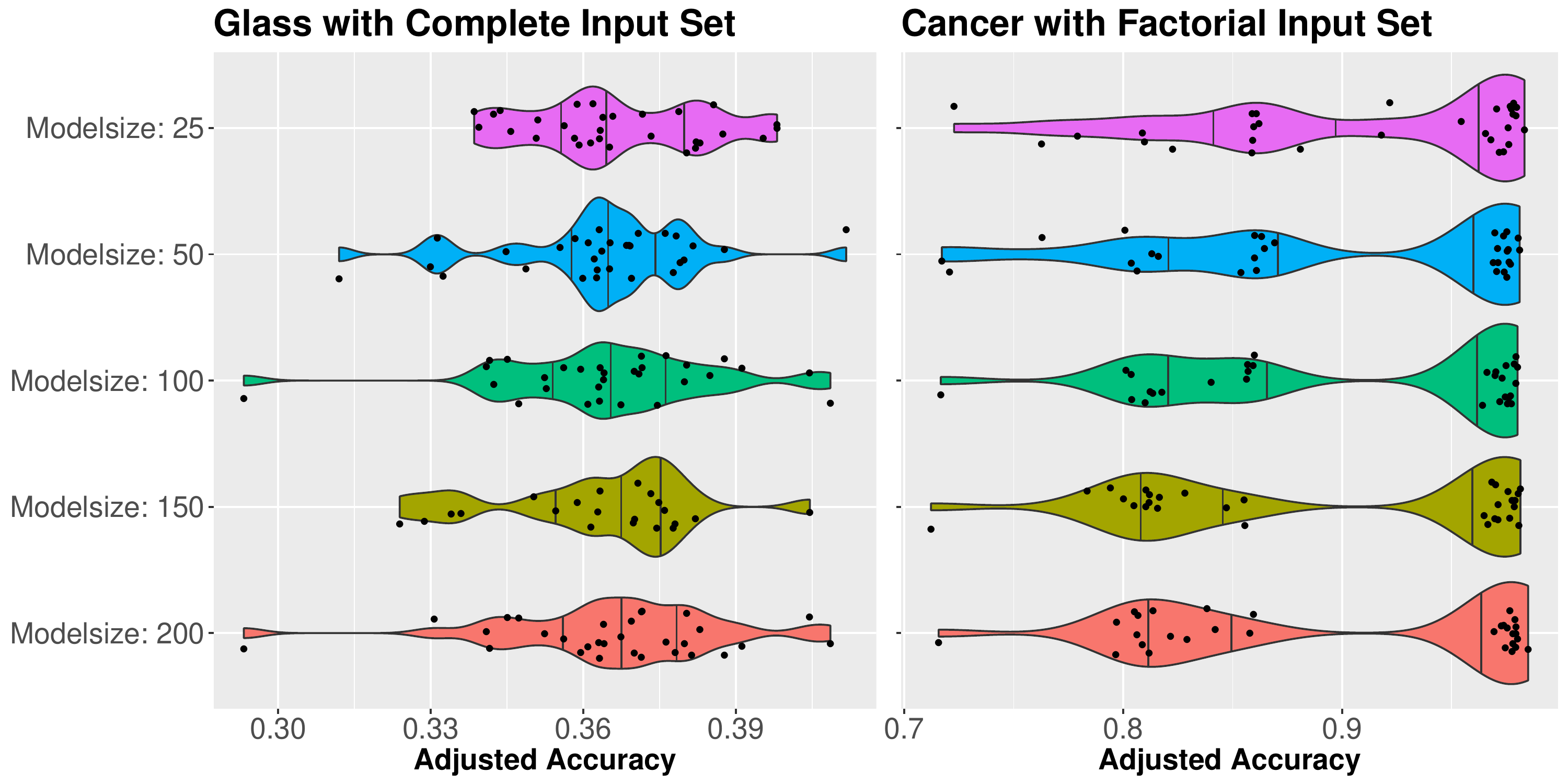}
\caption{Results with different surrogate model sizes during sequential optimization run for SMB.NE using PhD. Compared datasets are glass with complete input set for PhD, cancer with factorial input set for PhD.}
\label{fig:resultMsize}
\end{figure}
Fig.~\ref{fig:resultDOE} shows the results of the benchmarks with SMB-NE using PhD for different input sets created by experimental design techniques. 
For glass, the SMB-NE with the complete input set performed best and on one level with CGPANN with 100 times more real function evaluations. 
This indicates that SMB-NE is clearly able to improve the evaluation efficiency, if the (best possible) input data set is selected. All DOE input sets performed on an equal level similar to CGPANN with 10 times the evaluations. They are thus also able to significantly increase the evaluation efficiency, while utilizing an input set, which requires (nearly) no a priori information. 
The results for cancer firstly indicate that the underlying problem seems to include strong local optima, as we can identify certain clusters of different levels of the adjusted accuracy (around 0.8, 0.86, and 0.95). 
If we only consider only the best cluster, again we can identify that again the complete set performs best, together with CGPANN x100, while the DOE sets are again on one level with CGPANN x10. 

Fig. \ref{fig:resultMsize} shows the results of the benchmark with different surrogate model sizes during sequential optimization runs. 
In contrast to our expectations, the benchmarks show that the model sample size does not seem to have a significant impact on the performance. 
Even the small sample sizes perform on the same level.

\subsubsection{Discussion} 
The second set of experiments significantly shows two features of SMB-NE using the PhD distance measure. As expected, the choice of input set has a strong influence on the algorithm performance. The different designs and input sample sizes show a similar performance, which is an unexpected finding, as we would have anticipated that more samples and a larger design would lead to a more precise representational distance for the complete dataset and thus an improved performance. 
As the complete dataset shows the best performance, a representatively measured dataset of the original task seems to be the best choice, if initially available or producible by experiments. 
For the surrogate model size, we anticipated that the large model, which has more information, would be beneficial to the search process, but the smaller, more local models also worked well. 
Thus, it can be held on a feasible level which is fast to compute. 

\newpage
\section{Conclusion and Outlook} \label{sec:con}
In this work, we proposed a new surrogate-based approach for the task of NeuroEvolution. 
Further, we investigated the influence of different distance measures for constructing the surrogate during the optimization process. 
We have shown that SMBO is a valuable extension for CGPANN, which is able to improve the NeuroEvolution of ANNs in case of small evaluation budgets. 
Regarding research question 1, we can conclude that SMB-NE with phenotypic distance kernels shows significantly better results than basic CGPANN.
Utilizing the PhD distance measure with a perfect input set, we were able to reach an evaluation efficiency which is on one level with CGPANN with 100 times more function evaluations.
Further, the comparison indicates that SMB-NE utilizing genotypic distances does not provide significant performance increases. 
This fact can be explained by the low correlation of changes in the genotype to the resulting ANN behavior. 
Regarding research question 2, the experiments have shown that the choice of input sets for computing the PhD distance measure has a significant influence on
the algorithm performance. 
Input sets created by DOE methods show a reduced performance in comparison to a perfect input set.
Finally, we can conclude that the PhD distance measure, which is insensitive to the ANN size, is very promising.
Currently, we are working on applying SMB-NE with the PhD distance measure to reinforcement learning tasks, 
which pose additional challenges with regards to the choice of the input set.

\noindent
{\footnotesize {\textbf{Acknowledgements:}} 

This work is supported by the German Federal Ministry of Education and Research in the funding program Forschung an Fachhochschulen under the grant number 13FH007IB6.

\bibliographystyle{splncs03}
\bibliography{Stor18cnew}  

\end{document}